\documentclass[10pt,twocolumn,letterpaper]{article}

\usepackage[pagenumbers]{wacv} % To force page numbers, e.g. for an arXiv version

\definecolor{wacvblue}{rgb}{0.21,0.49,0.74}
\usepackage[pagebackref,breaklinks,colorlinks,allcolors=wacvblue]{hyperref}

\usepackage{multirow}

\def\wacvPaperID{3518} % *** Enter the WACV Paper ID here
\def\confName{WACV}
\def\confYear{2027}

\title{NOVA: Normal-Side Modeling for Training-Free Zero-Shot Video Anomaly Detection}

\author{Wei-Chih Yin, Yun-Ching Kao, Cheng-Kuan Lin, Yu-Chee Tseng\\
Department of Computer Science, National Yang Ming Chiao Tung University, Hsinchu, Taiwan\\
{\tt\small eric123602@gmail.com, yunching.cs14@nycu.edu.tw, cklin@cs.nycu.edu.tw, yctseng@cs.nycu.edu.tw}
}

\usepackage{listings}
\makeatletter
\@ifundefined{nolinenumbers}{\newcommand{\nolinenumbers}{}}{}
\@ifundefined{linenumbers}{\newcommand{\linenumbers}{}}{}
\lst@AddToHook{Init}{\nolinenumbers}
\lst@AddToHook{DeInit}{\linenumbers}
\makeatother
\begin{document}
\maketitle
% !TEX root = ../main.tex
\begin{abstract}
Training-free zero-shot video anomaly detection (ZS-VAD) leverages vision-language models (VLMs) to localize anomaly instances from a predefined anomaly vocabulary, without providing any video.
Existing CLIP-based methods often emphasize \textbf{anomaly-side semantics}, while the competing \textbf{normality side} remains less carefully formulated. We identify two key limitations in existing solutions: (i) \textbf{blurred decision boundary}: normal prompts may contain ambiguous verbs, such as ``running'', that are semantically close to anomalies, reducing normal and abnormal separation in the VLM embedding space; and (ii) \textbf{modality gap}: poor alignment between features of textual normal anchors and visual frames.
We propose \textbf{NOVA}, a training-free ZS-VAD framework that strengthens the normal side at both linguistic and visual levels. NOVA introduces \textbf{Normality-Aware Prompt Construction (NA)}, which excludes anomaly-adjacent verbs and biases normal descriptions toward static, low-motion scenes. To overcome the text-vision modality gap, NOVA constructs a \textbf{Visual Normality Anchor (VNA)} , which creates a weighted visual normal anchor from the initial frames of each test video, providing a video-specific normal reference without task-specific training or annotations.
%providing a per-video normal reference without requiring extra annotations or calibration. 
NOVA achieves 89.86\% AUC on UCF-Crime and 95.07\% AUC and 84.82\% AP on XD-Violence, reaching state-of-the-art performance among comparable training-free zero-shot methods.
\end{abstract}

% !TEX root = ../main.tex
\section{Introduction}
\label{sec:intro}

\textit{Video Anomaly Detection (VAD)} is essential in intelligent surveillance, aiming to localize rare and semantically diverse abnormal events, such as fighting, robbery, arson, and explosions, from long and untrimmed surveillance videos. \textit{Supervised and weakly supervised methods}~\cite{sultani2018real,tian2021weakly,cliptsa2022,wu2024vadclip} have achieved substantial progress in closed-set settings, but they rely on collecting and annotating anomalous samples from the target domain. This requirement is inherently restrictive: anomalies are rare, open-ended, and difficult to enumerate, making such methods less reliable when deployed in new scenes.

Vision-language models (VLMs), such as CLIP~\cite{radford2021learning}, offer a natural basis for \textit{zero-shot VAD (ZS-VAD)} by aligning visual frames and textual descriptions in a shared embedding space. By comparing frame-level visual embeddings against competing abnormal and normal text prompts, ZS-VAD can score anomalies that were never used for task-specific training. Recent methods enrich textual representations with LLM-generated descriptions, either as category-level prompts or as an offline pseudo-scene memory~\cite{dong2024clip,lee2025flashback}.

Despite these advances, existing CLIP-based ZS-VAD methods often place greater emphasis on enriching \textit{anomaly-side semantics}. However, anomalies are inherently open-ended and difficult to describe exhaustively. In prompt-contrastive ZS-VAD, anomaly scores are determined by the competition between anomaly and normal prompts rather than anomaly prompts alone. As a result, the normal side is not merely a background reference but directly shapes the decision boundary.
Nevertheless, \textit{normality semantics} remains underexplored in existing training-free ZS-VAD methods. Normal prompts are often represented by generic descriptions, despite normal events being more constrained by scene layout, object configuration, and video-specific appearance. This mismatch motivates us to revisit the role of the normal side in training-free ZS-VAD.

We identify two key limitations of current training-free ZS-VAD frameworks from the perspective of normal-side modeling.
The first limitation is the \textbf{blurred decision boundary} in LLM-generated normal descriptions. Although existing methods distinguish normal and abnormal prompts using separate templates or embedding-level repulsion, semantically ambiguous verbs may still appear in normal descriptions. For example, actions such as ``running'' or ``chasing'' may describe normal activities in some contexts, yet they are also associated with violent or suspicious events. Consequently, the semantic distinction between normal and abnormal prompts becomes less clear.
The second limitation is the inherent \textbf{modality gap} between textual and visual embeddings, which persists even with well-designed normal prompts~\cite{liang2022mindgap,song2025reducing}. A textual normal anchor does not necessarily lie closer to normal video frames than an abnormal textual description does. Furthermore, a fixed prompt bank shared across all videos cannot account for the appearance differences among individual videos. This discrepancy originates from the gap between textual and visual representations, rather than prompt wording alone.

To address these limitations, we propose \textbf{NOVA} (\textbf{NO}rmal-side Modeling for ZS-\textbf{VA}D), a training-free framework that explicitly models the normal side from both linguistic and visual perspectives.
On the language side, NOVA introduces \textbf{Normality-Aware Prompt Construction (NA)}, which refines LLM-generated normal descriptions to improve the semantic distinction between normal and abnormal prompts.
On the visual side, NOVA introduces the \textbf{Visual Normality Anchor (VNA)}, a per-test-video normal reference constructed directly in the visual embedding space. By estimating the normal reference from the test video itself, VNA alleviates the discrepancy between textual normal descriptions and visual representations while adapting to each video.
In addition, NOVA incorporates a lightweight temporal module, \textbf{Motion-Aware Stabilization (MAS)}, to improve frame-level temporal stability. The resulting framework remains fully training-free, using a frozen vision-language encoder, a pre-constructed prompt bank, and test-time visual anchors.

Our contributions are fourfold. First, we propose \textbf{NOVA}, a training-free framework for zero-shot video anomaly detection that explicitly treats normality through complementary linguistic and visual modeling.
Second, NOVA introduces \textbf{Normality-Aware Prompt Construction (NA)} to reduce semantic ambiguity in normal descriptions and the \textbf{Visual Normality Anchor (VNA)} to construct a per-video visual normal reference directly from the test video, and further incorporates a lightweight \textbf{Motion-Aware Stabilization (MAS)} module for temporal refinement.
Third, through controlled ablation studies and embedding-space analyses, we demonstrate the importance of explicit normal-side modeling in training-free ZS-VAD and provide empirical evidence for the effectiveness of the proposed design.
Finally, in a zero-shot setting, NOVA achieves state-of-the-art performance among comparable training-free ZS-VAD methods, reaching 95.07\% AUC on XD-Violence and 89.86\% AUC on UCF-Crime.

% !TEX root = ../main.tex
\section{Related Work}
\label{sec:related}

\subsection{Video Anomaly Detection}
\label{sec:vad}

VAD aims to localize rare abnormal events in long, untrimmed videos.
Existing VAD methods rely on frame-level annotations, video-level labels,
or unlabeled target-domain videos to learn anomaly or normality
representations~\cite{ramachandra2020survey,sultani2018real,tian2021weakly,cliptsa2022,wu2024vadclip,zaheer2022generative}.
Despite their success, these approaches require task-specific data and learn dataset-specific notions of normality and abnormality.
Recent open-vocabulary methods~\cite{li2025anomize,liu2026language} relax the closed-set assumption but still require task-specific training or adaptation.

\subsection{Zero-Shot Video Anomaly Detection}
\label{sec:zsvad}

Training-free ZS-VAD aims to detect previously unseen anomalies at inference time using frozen VLMs or MLLMs without collecting target-domain training data~\cite{zanella2024harnessing,lee2025flashback,askhint2025,shao2025eventvad,ahn2026anyanomaly}. Existing methods can be categorized into \textit{prompt-contrastive} and \textit{reasoning-based} approaches.
Prompt-contrastive methods estimate anomaly scores by comparing visual representations against competing normal and abnormal textual descriptions. 
For image anomaly detection, WinCLIP~\cite{jeong2023winclip} compares image embeddings with handcrafted normal and abnormal prompts for zero-shot anomaly classification and localization. It was later adopted for video anomaly detection. VadCLIP~\cite{wu2024vadclip} aligns textual event categories with video representations under weak supervision, while \cite{dong2024clip} introduces LLM-generated normal and abnormal descriptions but still trains learnable prompts and temporal modules. Flashback~\cite{lee2025flashback} removes task-specific training by constructing an offline pseudo-scene memory while preserving normal--abnormal prompt competition during inference. NOVA follows this prompt-contrastive formulation but differs in its treatment of normality: unlike prior approaches whose normal references remain text-derived, NOVA explicitly disambiguates normal descriptions and constructs a per-video visual normality anchor directly from the test video.

Reasoning-based training-free methods instead use frozen VLMs or MLLMs to infer anomaly scores directly. LAVAD~\cite{zanella2024harnessing} converts VLM-generated scene descriptions into anomaly scores through LLM reasoning, while Cerberus~\cite{zheng2025cerberus}, VADTree~\cite{li2026vadtree}, ASK-Hint~\cite{askhint2025}, AnyAnomaly~\cite{ahn2026anyanomaly}, EventVAD~\cite{shao2025eventvad}, and PANDA~\cite{Yang2025PANDATG} explore rule-based or hierarchical reasoning, structured prompting, and multimodal inference. 
LAVIDA~\cite{lavida2026} also uses an MLLM, but trains on pseudo-anomalies synthesized from external segmentation data and therefore falls outside the strictly training-free setting. Compared with these direct-reasoning approaches, NOVA retains a lightweight prompt-contrastive scorer and focuses on strengthening normal-side representations.

% NOVA instead retains a lightweight prompt-contrastive scorer and improves the normal side used in that competition.

% \noindent\textbf{Differences from text-based ZS-VAD methods.}

\subsection{Normality Modeling}
\label{sec:normalmodeling}

Normality plays a fundamental role in anomaly detection. In \textit{training-based} VAD, normality is learned from target-domain videos. One-class VAD methods learn normality from normal-only videos and regard deviations from the learned normal patterns as anomalies~\cite{wang2019gods}. Weakly supervised VAD additionally uses video-level labels to learn normality and regularize anomaly scoring~\cite{sultani2018real,tian2021weakly}. 
% Although these settings differ in the degree of supervision, both rely on target-domain training data to establish normality.
In \textit{training-free} ZS-VAD, normality is instead specified without collecting target-domain training videos, typically through textual prompts or text-only memories constructed before inference~\cite{lee2025flashback}. Cerberus~\cite{zheng2025cerberus} derives scene-specific normal behavioral rules from sample normal videos during an offline induction phase; it is thus data-adaptive rather than strictly target-data-free. 
Consequently, effective normal representations must capture normality while remaining distinguishable from anomalies.
INP-Former~\cite{luo2025exploring} extracts intrinsic normal prototypes directly from each test image, but requires training.
This suggests that test-instance-specific visual normality can complement text-based normal representations. NOVA similarly constructs a video-specific visual normality anchor from the test input, but requires neither normal training videos nor task-specific training.

\subsection{Prompt Design and Modality Gap}
\label{sec:promptdesign}

Prompt design plays an important role in adapting vision-language models (VLMs) to downstream tasks. General prompt learning and prompt generation methods have shown that textual context can substantially influence visual recognition. 
CoOp~\cite{zhou2022coop} learns task-adaptive context vectors from labeled data, whereas CuPL~\cite{pratt2023cupl} leverages LLM-generated textual descriptions to improve zero-shot classification. In anomaly detection, prompt design is particularly important because anomaly scores are determined by the contrast between normal and abnormal descriptions rather than a single class label. Image anomaly detection methods such as WinCLIP~\cite{jeong2023winclip} and AnomalyCLIP~\cite{zhou2024anomalyclip} represent normal and abnormal states through handcrafted or learned prompts, while video anomaly detection methods adopt related prompt-based representations for VAD~\cite{dong2024clip,lee2025flashback}.

However, prompt engineering alone does not fully address the \textit{modality gap} between textual and visual representations. Prior studies have shown that contrastive VLMs may embed text and image features into different regions of a shared feature space despite being trained for cross-modal alignment~\cite{liang2022mindgap,song2025reducing}. For video anomaly detection, this implies that even semantically appropriate textual normal prompts need \textit{not} lie close to normal video features in the shared embedding space. Consequently, a semantically reasonable prompt bank may still be suboptimal.

% !TEX root = ../main.tex
\section{Method}
\label{sec:method}

\subsection{Problem Definition and Framework}
\label{sec:problem_formulation}

Let $V=\{f_t\}_{t=1}^{T}$ denote a test video with $T$ frames.
Given $V$, a coarse footage descriptor $d$ (e.g., ``surveillance video''),
and a predefined anomaly vocabulary $\mathcal{C}$ (e.g.,
\{``fighting'', ``shooting''\}), NOVA estimates frame-level anomaly
scores $s_t\in[0,1]$, where higher values indicate a higher likelihood
of anomaly.
The descriptor $d$ and vocabulary $\mathcal{C}$ are used to
construct the prompt bank before inference.
NOVA requires no task-specific training or target-domain adaptation,
and frame-level ground truth $y_t\in\{0,1\}$ remains unavailable
throughout.

\begin{figure*}[ht]
\centering
\includegraphics[width=\textwidth]{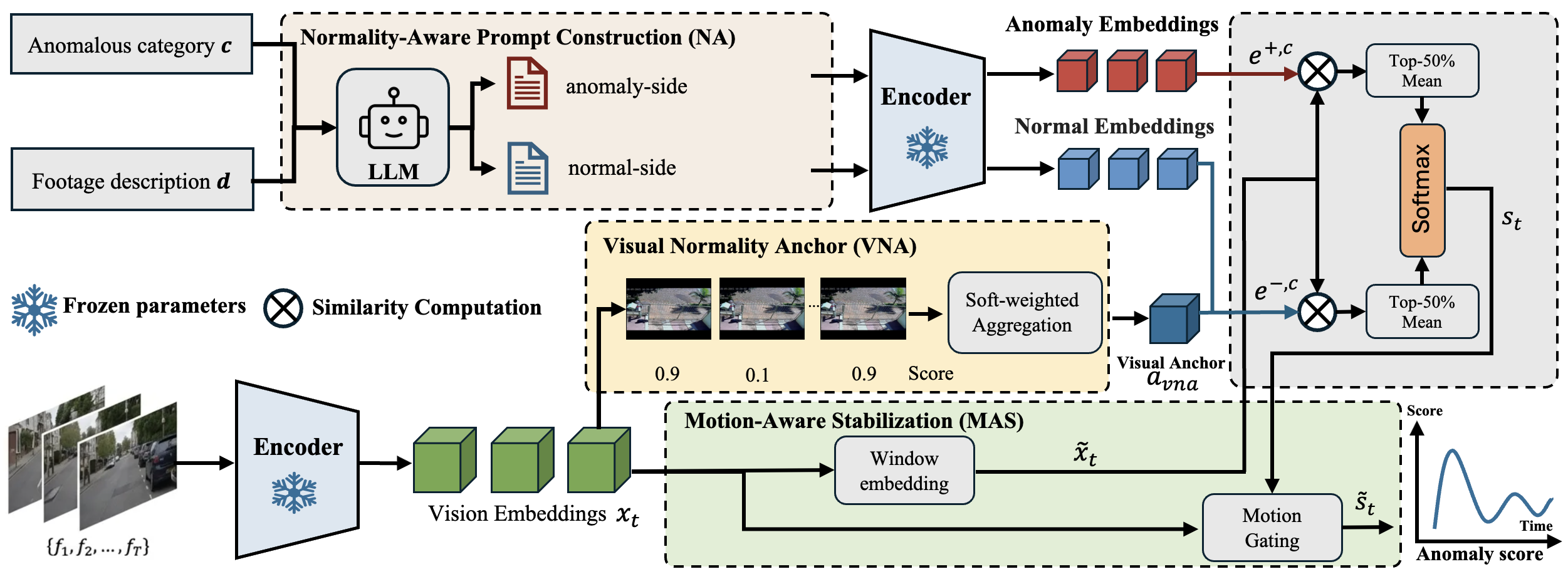}
\caption{The NOVA framework. NOVA uses a frozen vision-language encoder and a pre-constructed prompt bank, and constructs a video-specific visual normality anchor at test time. The framework comprises four key components: Normality-Aware Prompt Construction (NA), prompt-contrastive anomaly scoring, a per-video Visual Normality Anchor (VNA), and Motion-Aware Stabilization (MAS).}
\label{fig:framework}
\end{figure*}

As illustrated in Fig.~\ref{fig:framework}, NOVA comprises four
components: \textbf{Normality-Aware Prompt Construction (NA)}
constructs anomaly and disambiguated normal descriptions
(Sec.~\ref{sec:prompts}); \textbf{Prompt-Contrastive Anomaly Scoring}
performs normal--abnormal competition using a frozen vision-language
encoder (Sec.~\ref{sec:scoring}); \textbf{Visual Normality Anchor (VNA)}
introduces a per-video visual normal reference
(Sec.~\ref{sec:vna}); and \textbf{Motion-Aware Stabilization (MAS)}
provides lightweight temporal refinement
(Sec.~\ref{sec:MAS}).
The entire framework is training-free and requires no parameter
optimization.

%NOVA consists of four complementary modules (Fig.~\ref{fig:framework}).  \textbf{Normality-Aware Prompt Construction (NA)} first builds anomaly descriptions and disambiguated normal descriptions for each anomaly category $c\in\mathcal{C}$ (Sec.~\ref{sec:prompts}). \textbf{Prompt-contrastive Anomaly Scoring} then uses a frozen vision-language encoder to map frames and prompts from both sides into a shared embedding space, producing category-conditioned scores through anomaly/normal competition (Sec.~\ref{sec:scoring}). On top of this backbone, VNA incorporates early visual normal anchors from the test video itself into the normal-side pool, alleviating the modality gap of textual normal prompts (Sec.~\ref{sec:vna}); MAS further introduces lightweight temporal enhancement at both the embedding and score levels (Sec.~\ref{sec:MAS}). The entire framework trains no parameters.

%Normality-Aware 
\subsection{Normality-Aware Prompt Construction (NA)}
\label{sec:prompts}

All prompt banks are constructed using an LLM, but no LLM is invoked during inference.

For each anomaly category $c\in\mathcal{C}$, our goal is to construct a category-conditioned prompt bank
$
\mathcal{B}_c=\left\{
\mathcal{P}_c =\{p_i^c\}_{i=1}^{M},
\mathcal{N}_c=\{n_j^c\}_{j=1}^{M}
\right\}$,
where $\mathcal{P}_c$ contains anomaly-side descriptions, $\mathcal{N}_c$ contains normal-side descriptions, and $M$ is their cardinality.
Each anomaly description $p_i^c \in \mathcal{P}_c$ is expected to cover a subject, an object, and an observable atomic event corresponding to $c$.
Each normal description $n_j^c \in \mathcal{N}_c$, in contrast, should provide \textit{safe, low-motion, and non-threatening counterexamples} within the same context. A key requirement is that the normal side should not merely describe generic normality, but must avoid \textit{visually similar normal behaviors} related to $c$; otherwise, the normal descriptions may lie close to the anomaly side in the textual embedding space and weaken the subsequent competitive scoring. 

% The construction pipeline below realizes the normal-side constraints that enforce this requirement; we abbreviate this normality-aware prompt construction as \textbf{NA}.

For each category $c$, the prompt bank $\mathcal{B}_c$ is constructed in four steps. Step 4 is potentially repeated multiple times.

\textbf{Step 1. Confusing verb mining.}
Given $c$ and the footage descriptor $d$, an LLM is invoked to mine a set of category-conditioned yet \textit{confusing} normal actions,
$\mathcal{V}_c = \mathrm{LLM}_{\mathrm{verb}}(c, d)$, where each element is a short phrase consisting of 1–3 words that describes a semantically normal action visually similar to $c$.  A single query asks the LLM for 5–8 such actions. For example, $\mathcal{V}_{\mathrm{fighting}}$ may include ``sparring'', ``play wrestling'', and ``horseplay''.

\textbf{Step 2. Calm-anchor generation.} 
Conditioned only on $d$, an LLM is invoked to generate a fixed set of calm anchors,
$\mathcal{A} = \mathrm{LLM}_{\mathrm{anchor}}(d) = \{a_1, \dots, a_K\}$,
where each $a_k$ is a low-motion, non-threatening baseline scene of the footage domain, such as ``an empty corridor''. As $\mathcal{A}$ does not depend on $c$, it is generated once per prompt-bank run and shared across all categories, providing a stable normal reference for common background or establishing scenes in the footage domain.

\textbf{Step 3. Prompt bank generation.} 
Using a single instruction, an LLM is invoked to produce the anomaly set and an initial normal set,
$
(\mathcal{P}_c, \mathcal{N}_c) = \mathrm{LLM}_{\mathrm{gen}}(c, \mathcal{V}_c, \mathcal{A}, d)$,
where $|\mathcal{P}_c| = |\mathcal{N}_c| = M$. The anomaly set $\mathcal{P}_c$ is produced once and fixed, whereas the normal set $\mathcal{N}_c$ may be further revised in step 4.
Following typical anomaly-driven work, $\mathcal{P}_c$ directly describes the category $c$ itself 
(e.g., ``Two people exchanging rapid punches to each other's faces in a hallway'' for category ``fighting''). 
Enforcing $M>K$, $\mathcal{N}_c$ consists of $K$ stable establishing-scene descriptions generated from the calm-anchor reference $\mathcal{A}$ (e.g., ``A quiet empty hallway under fluorescent lights with closed doors and no people'') and $M-K$ ordinary normal descriptions (e.g., ``Two people talking calmly beside a doorway with relaxed posture''). 
The generation instruction explicitly \textit{prohibits} terms in the confusing set $\mathcal{V}_c$ from appearing in the normal descriptions. This constraint prevents semantically ambiguous verbs, such as ``sparring'', from appearing in normality descriptions.
Because the anchor set $\mathcal{A}$ provides examples rather than verbatim templates, the anchor-derived descriptions differ across categories instead of repeating one fixed set of sentences.

\textbf{Step 4. Geometry-gated refinement.} 
After step 3, some descriptions in the normal set $\mathcal{N}_c$ may still lean toward the anomaly side in the textual embedding space. This step geometrically detects and rewrites such residual cases.
Assuming a pretrained text encoder $\psi(\cdot)$, we first compute the normalized centroid vector of the anomaly bank,
\begin{equation}
\boldsymbol{\mu}_c^{+}=\mathrm{Normalize}\!\left(\frac{1}{M}
\sum\nolimits_{p_i^c \in \mathcal{P}_c} 
\psi(T^{+}, p_i^c)\right).
\end{equation}
From the centroid, we compute the ambiguity score $g_j^{c}$ of each normal description $n_j^{c} \in \mathcal{N}_c$ via cosine similarity,
\begin{equation}
g_j^{c}=\psi(T^{-}, n_j^{c})^{\top}\boldsymbol{\mu}_c^{+},
\end{equation}
where a higher value indicates that the normal description is closer to the anomaly side.
$T^+$ and $T^-$ are predefined text prefixes for embedding.
The goal of refinement is to rewrite description $n_j^{c}$ to remove potential ambiguity.
Specifically, $n_j^{c}$ is marked as an \textit{ambiguous normal} if its $g_j^{c}>\theta$, in which case we invoke an LLM for revision:
$\hat{n}_j^{c}=\mathrm{LLM}_{\mathrm{ref}}(n_j^{c}, c)$ and similarly evaluate its ambiguity score
$\hat{g}_j^{c}$.
The revision replaces $n_j^{c}$ with $\hat{n}_j^{c}$ only if the ambiguity is reduced, i.e., $\hat{g}_j^{c}<g_j^{c}$; otherwise the original $n_j^{c}$ is kept.
The above scoring, marking, and rewriting are then repeated over the current $\mathcal{N}_c$, until the ambiguity scores of all its descriptions drop below $\theta$, until a round accepts no revision, or until a predefined number $R$ of rounds is reached. The resulting $\mathcal{N}_c$, together with $\mathcal{P}_c$, forms the prompt bank $\mathcal{B}_c$.
Values of $T^+$, $T^-$, $\theta$, and $R$ are reported in Sec.~\ref{sec:datasets}.

The dataset-specific inputs include only the anomaly vocabulary $\mathcal{C}$, footage descriptor $d$, and fixed domain-specific
role guidance and examples used for prompt generation.
These specifications are defined once per dataset and held fixed across categories, seeds, and prompt-generation runs.
For UCF-Crime, $d=$~``surveillance footage'', whereas for XD-Violence, $d=$~``movie or online video footage''.

% \red{
% \begin{equation}
% n_j^{c,(t+1)}=
% \begin{cases}
% \hat{n}_j^{c,(t)}, & g_j^{c,(t)}>\theta \ \text{and}\ \hat{g}_j^{c,(t)}<g_j^{c,(t)},\\[2pt]
% n_j^{c,(t)}, & \text{otherwise.}
% \end{cases}
% \end{equation}
% }

% \red{prefix templates $\mathcal{T}_A(\cdot)$, $\mathcal{T}_N(\cdot)$ as inference-time scoring (Sec.~\ref{sec:scoring}). 
% }

\subsection{Prompt-Contrastive Anomaly Scoring}
\label{sec:scoring}

Prompt-contrastive scoring adopts the \textit{Repulsive Prompting (RP)} and \textit{Scaled Anomaly Penalization (SAP)} principles of Flashback~\cite{lee2025flashback}, and applies them to category-conditioned competition between anomaly and normal prompts.
%We first establish the prompt-contrastive scoring backbone upon which VNA and MAS are built.
Based on the same text encoder $\psi(\cdot)$ and the prompt bank $(\mathcal{P}_c, \mathcal{N}_c)$, we compute two L2-normalized textual embedding pools as
$\mathbf{e}_{i}^{+,c}=\psi(T^{+}, p_i^c)$ and $\mathbf{e}_{j}^{-,c}=\psi(T^{-}, n_j^c)$,
where
$p_i^c \in \mathcal{P}_c$ and
$n_j^c \in \mathcal{N}_c$.
% Assuming a pretrained visual encoder $\phi(\cdot)$, we also compute an L2-normalized visual embedding for each frame $f_t$,
%Assuming a frozen vision-language encoder whose text tower is the $\psi(\cdot)$ above and whose visual tower is $\phi(\cdot)$, so that textual and visual embeddings live in one shared space, we also compute an L2-normalized visual embedding for each frame $f_t$,
%$\mathbf{x}_t=\phi(f_t)$.
Using the corresponding visual encoder $\phi(\cdot)$ from the same frozen vision-language model, each frame is represented by the L2-normalized visual embedding
$\mathbf{x}_t=\phi(f_t)$.
%We adopt Repulsive Prompting and SAP-Softmax from Flashback~\cite{lee2025flashback} as the scoring rule. 
% Anomaly and normal descriptions are wrapped with different prefix templates $T^{+}$ and $T^{-}$, such as ``Anomalous scene:'' and ``Normal scene:'', forming two textual embedding pools:
The similarities between $\mathbf{x}_t$ and the anomaly-side and normal-side embeddings are computed respectively as:
\begin{equation}
r_i^{+,c}=\mathbf{x}_t^\top\mathbf{e}_{i}^{+,c},\qquad
r_j^{-,c}=\mathbf{x}_t^\top\mathbf{e}_{j}^{-,c}.
\end{equation}
To reduce the noise of individual prompts, we apply Top-50\% mean aggregation to estimate the similarities of $f_t$ to both sides:
\begin{equation}
\begin{aligned}
u_t^{+,c} &= \mathrm{TopKMean}_{50\%}(\{r_i^{+,c}\}),\\
u_t^{-,c} &= \mathrm{TopKMean}_{50\%}(\{r_j^{-,c}\}).
\end{aligned}
\end{equation}

Following the SAP principle~\cite{lee2025flashback}, we compensate for the asymmetry between anomaly and normal prompts by down-weighting the anomaly similarity with a scaling factor $\alpha<1$. 
The $c$-conditioned anomaly score is computed by binary Softmax, written in the equivalent sigmoid form below:
\begin{equation}
s_t^c =
\operatorname{sigmoid}\!\left(
\frac{\alpha u_t^{+,c}-u_t^{-,c}}{\tau}
\right).
\end{equation}
The same $\alpha$ and $\tau$ are used across all datasets without target-specific tuning. Their values are reported in Sec.~\ref{sec:datasets}.

The final frame-level anomaly score is obtained as $
s_t=\max_{c\in\mathcal{C}} s_t^c$.
The scoring backbone above serves as the basis for the following two modules. VNA (Sec.~\ref{sec:vna}) strengthens the normal-side representation, while MAS (Sec.~\ref{sec:MAS}) further improves temporal stability.

% without altering the underlying scoring formulation.

%\paragraph{Top-$K$ similarity aggregation.}
%The temporal replacement of $\mathbf{x}_t$ under MAS is described in Sec.~\ref{sec:MAS}. 

\subsection{Visual Normality Anchor (VNA)}
\label{sec:vna}

Textual normal prompts provide semantic guidance, but because they are fixed across videos, they cannot capture the scene-specific appearance of an individual video. Moreover, even semantically appropriate textual prompts may remain geometrically separated from visual frame embeddings because of the \textit{modality gap}~\cite{liang2022mindgap,song2025reducing}. To complement these textual references, VNA constructs a \textit{per-video} visual normality anchor directly from the test video.

\textbf{Warm-Up Normality Prior.}
After temporal sampling, we use the first $N_{\mathrm{warm}}$ sampled visual embeddings as a lightweight estimate of the video's normal visual state without requiring annotations. When MAS is enabled, its windowed embedding aggregation is applied before temporal sampling and VNA estimation. Since the early-video assumption may not hold for every video, VNA incorporates a contamination-aware weighting scheme.
%Surveillance anomalies are typically sparse and temporally localized events, while normal activities dominate most video segments~\cite{sultani2018real}. Surveillance anomalies are typically sparse and temporally localized, while normal activities dominate most video segments~\cite{sultani2018real}. We therefore use the first $N_{\mathrm{warm}}$ frames as a lightweight estimate of the video's normal visual state without requiring annotations. Since this assumption may not hold for every video, VNA further incorporates a contamination-aware weighting scheme.
%This prior is probabilistic and may fail on individual videos, which motivates the robust estimation below.

\textbf{Anomalous Contamination Avoidance.}
Directly averaging the warm-up embeddings is vulnerable to anomalous contamination: even a few anomalous embeddings can shift the visual prototype toward the anomaly side. To reduce this effect, we first compute a preliminary anomaly score $\hat{s}_t$ for each warm-up embedding using the scoring rule in Sec.~\ref{sec:scoring}. 
The resulting scores are normalized by:
\begin{equation}
w_t=
\frac{\exp(-\hat{s}_t/\gamma)}
{\sum_{q=1}^{N_{\mathrm{warm}}}\exp(-\hat{s}_q/\gamma)},
\end{equation}
where $\gamma$ is the anchor temperature that controls how strongly the weights concentrate on the lowest-scoring frames.
The visual normality anchor is then estimated as
\begin{equation}
\mathbf{a}_{\mathrm{VNA}}=
\mathrm{Normalize}\left(
\sum\nolimits_{t=1}^{N_{\mathrm{warm}}}w_t\mathbf{x}_t
\right).
\end{equation}

During anomaly scoring, the normal embedding pool is extended from
$\{\mathbf{e}_{j}^{-,c}\mid j=1,2,\ldots,M\}$
to
$\{\mathbf{e}_{j}^{-,c}\mid j=1,2,\ldots,M\}\cup\{\mathbf{a}_{\mathrm{VNA}}\}$.
The same visual anchor is shared across all anomaly categories, providing a video-specific normal reference in addition to the textual normal prompts.
We use the same $N_{\mathrm{warm}}$ and anchor temperature $\gamma$ for both datasets (values reported in Sec.~\ref{sec:datasets}).

%This design allows the visual-modality normal anchor to \textbf{participate in} the textual-modality normal pool and jointly compete against the textual-modality anomaly pool.
%Concretely, the anchor \textbf{extends the normality boundary}, reducing the likelihood of false positives when a normal visual frame lies too close to anomaly descriptions.

%change name

%\subsection{\red{Windowed Embedding and Motion Modulation}}

\subsection{Motion-Aware Stabilization (MAS)}
\label{sec:MAS}

Frame-level scoring is susceptible to fluctuations and may disrupt event continuity.
This module leverages motion information to stabilize the anomaly scores. It combines two ideas: windowed embedding and motion gate.

\textbf{Windowed Embedding.}
Given the visual embedding sequence 
$\{\mathbf{x}_t\}_{t=1}^{T}$, we employ moving average with a sliding window of width $W$ (boundary cases omitted for simplicity of presentation),
\begin{equation}
\tilde{\mathbf{x}}_t=
\mathrm{Normalize}\left(
\frac{1}{W}
\sum\nolimits_{k=-(W-1)/2}^{(W-1)/2}\mathbf{x}_{t+k}
\right).
\end{equation}
This windowed averaging stabilizes the frame-level embeddings. We consistently use $W=5$ in all experiments. When this module is enabled, the visual embedding $\mathbf{x}_t$ in Sec.~\ref{sec:scoring} is replaced by $\tilde{\mathbf{x}}_t$ to improve anomaly scoring.

\textbf{Motion Gating.}
Anomalous events are often accompanied by observable motion, whereas low-motion segments are a common source of false positives. We estimate the local motion strength of frame $f_t$ using the cosine distance between its neighboring original frame embeddings,
\begin{equation}
m_t=1-\mathbf{x}_{t-1}^{\top}\mathbf{x}_{t+1}.
\end{equation}
Note that $\mathbf{x}_{t-1}$ and $\mathbf{x}_{t+1}$ denote the L2-normalized frame embeddings before windowed aggregation.

A soft gate derived from the motion strength is then applied to refine the anomaly score of frame $f_t$:
\begin{equation}
\label{eq:gating}
\bar{s}_t=s_t
\cdot\mathrm{sig}\left(\beta(m_t-\tau_m)\right),
\end{equation}
where $\mathrm{sig}(\cdot)$ denotes the logistic sigmoid, $\tau_m$ is the motion threshold, and $\beta$ is the gate slope; the latter two are fixed across datasets without target-specific tuning, and their values are reported in Sec.~\ref{sec:datasets}. Unlike a hard threshold, the soft gate does not set low-motion scores exactly to zero. It down-weights low-motion frames, while high-motion frames largely retain their pre-gate scores.
When motion gating is enabled, the anomaly score $s_t$ in Sec.~\ref{sec:scoring} is replaced by $\bar{s}_t$ before Gaussian smoothing.
The final scores are then smoothed by a Gaussian temporal filter of bandwidth $\sigma$, a standard post-processing step separate from the embedding- and motion-level operations of MAS.

\textit{Remark.} In implementation, the full-rate visual embeddings are first used to compute the motion signal, after which windowed aggregation is applied when MAS is enabled. The resulting embeddings are temporally sampled (1:16) for prompt-contrastive scoring. The sampled anomaly scores are then linearly interpolated to the original frame rate, followed by motion gating in Eq.~\ref{eq:gating} and Gaussian smoothing.

% !TEX root = ../main.tex
\section{Experiments}
\label{sec:experiments}

\subsection{Experimental Setup}
\label{sec:datasets}

\textbf{Datasets.} 
We evaluate NOVA on two standard VAD benchmarks. \textbf{UCF-Crime}~\cite{sultani2018real} contains 1,900 surveillance videos, including 290 test videos with frame-level binary annotations. The dataset covers 13 anomaly categories: \textit{Abuse}, \textit{Arrest}, \textit{Arson}, \textit{Assault}, \textit{Burglary}, \textit{Explosion}, \textit{Fighting}, \textit{Road Accident}, \textit{Robbery}, \textit{Shooting}, \textit{Shoplifting}, \textit{Stealing}, and \textit{Vandalism}.
\textbf{XD-Violence}~\cite{wu2020not} contains videos collected from YouTube and movies, covering six anomaly categories: \textit{Abuse}, \textit{Car Accident}, \textit{Explosion}, \textit{Fighting}, \textit{Riot}, and \textit{Shooting}. The test set includes videos from multiple source domains and provides frame-level binary annotations.
Following~\cite{zanella2024harnessing,ye2025vera,lee2025flashback}, we evaluate only on the official test sets to ensure fairness. Frame-level AUC-ROC is reported on UCF-Crime, while frame-level average precision (AP) and frame-level AUC-ROC are reported on XD-Violence.

\textbf{Implementation Details.}
We use GPT-5.4-mini-2026-03-17~\cite{openai2026gpt} (OpenAI API) to construct the prompt banks described in Sec.~\ref{sec:prompts}. The dataset-specific role guidance and examples used by the generator are fixed across categories, seeds, and NA variants; their full contents are provided in the supplementary material. We use the text prefixes $T^+=$``Anomalous scene: '' and $T^-=$``Normal scene: '' when embedding anomaly and normal descriptions, respectively. For each anomaly category, we generate $M=20$ anomaly descriptions and $M=20$ normal descriptions. Normal-prompt construction uses category-specific confusing actions together with $K=4$ shared calm anchors, followed by geometry-guided refinement with threshold $\theta=0.70$ for at most $R=3$ rounds. All prompt-construction parameters ($K$, $M$, $\theta$, and $R$) are fixed a priori and are not calibrated on the target dataset. To evaluate prompt robustness, we independently generate five prompt banks for each category and report the average performance.
All experiments are conducted on a single NVIDIA GeForce RTX 4080 SUPER. Videos are uniformly sampled at an interval of 16 frames, and the linear interpolation described in the remark of Sec.~\ref{sec:MAS} is applied to recover frame-level predictions. Following prior work, dataset-level AUC and AP are computed by pooling frame-level predictions and binary labels from all test videos.
We use PE-Core-L14-336~\cite{meta2025pecore} as the frozen vision-language backbone. Unless otherwise specified, the same hyperparameters are used for both datasets: SAP coefficient $\alpha=0.90$, temperature $\tau=0.03$, Gaussian smoothing parameter $\sigma=50.0$, warm-up length $N_{\mathrm{warm}}=32$, VNA temperature $\gamma=0.3$, MAS window size $W=5$, motion threshold $\tau_m=0.02$, and gate slope $\beta=20$.

\subsection{Comparison with State-of-the-Art}
\label{sec:comparison}

Table~\ref{tab:sota} compares NOVA with representative VAD methods under different training paradigms. Since LLM-generated prompt banks may vary across calls, NOVA's results are reported as the mean over five independently generated prompt banks, whereas the results of competing methods are taken from their respective papers.

Among training-free zero-shot methods, NOVA achieves the best performance on all three evaluation metrics. On UCF-Crime, NOVA obtains the highest frame-level AUC. On XD-Violence, NOVA achieves the highest AUC and AP, improving AP over Flashback~\cite{lee2025flashback} by \textbf{9.69} percentage points and AUC over VADTree~\cite{li2026vadtree} by \textbf{4.52} percentage points.
Despite requiring no target-domain training, fine-tuning, or calibration,
NOVA remains competitive with several weakly supervised methods and
outperforms the other training-free zero-shot methods included in the comparison.

\begin{table}[t]
  \centering
 \caption{
Comparison with VAD methods under different training paradigms.
\textbf{Bold} indicates the best result, while \underline{underline} indicates the best result within a paradigm.
``---'' denotes results not reported in the original paper.
$^{\dagger}$LAVIDA and $^{\dagger}$LaGoVAD use external segmentation data to synthesize pseudo-anomalies and are therefore not considered strictly training-free.
}
  \label{tab:sota}
  \resizebox{1.0\columnwidth}{!}{%
 \small
 \setlength{\tabcolsep}{4pt}
  
  \begin{tabular}{llccc}
    \toprule
    &     \multirow{2}{*}[-0.5ex]{%
\begin{tabular}{@{}l@{}}
Method
\end{tabular}} & UCF-Crime & \multicolumn{2}{c}{XD-Violence} \\
    \cmidrule(lr){3-3} \cmidrule(lr){4-5}
    &  & AUC (\%)$\uparrow$ & AP (\%)$\uparrow$ & AUC (\%)$\uparrow$ \\
    \midrule
    \multirow{6}{*}{%
\begin{tabular}{@{}l@{}}
{\textit{Weakly}}\\{\textit{supervised}}
\end{tabular}}
&   RareAnom~\cite{thakare2023rareanom} & 83.56 & --- & 79.89 \\
 &   VERA~\cite{ye2025vera}              & 86.55 & --- & \underline{88.26} \\
 &   CLIP-TSA~\cite{cliptsa2022}         & 87.58 & 82.19 & --- \\
 &   VadCLIP~\cite{wu2024vadclip}        & 88.02 & 84.51 & --- \\
 &   Dong~\cite{dong2024clip}            & 88.52 & --- & --- \\
 &   Holmes-VAD~\cite{zhang2024holmes}   & \underline{89.51} & \underline{\textbf{90.67}} & --- \\
    \midrule
    
   {\textit{One class}} 
    & GODS~\cite{wang2019gods}            & 70.46 & --- & --- \\
    \midrule

        \multirow{4}{*}{%
\begin{tabular}{@{}l@{}}
{\textit{Unsupervised}}
\end{tabular}}
   &
    GCL~\cite{zaheer2022generative}     & 71.04 & --- & --- \\
   & FPDM~\cite{yan2023feature}          & 74.70 & --- & --- \\
   & MULDE~\cite{micorek2024mulde}       & 78.50 & --- & --- \\
   & DyAnNet~\cite{thakare2023dyannet}   & \underline{84.50} & --- & --- \\
    \midrule

        \multirow{2}{*}{%
\begin{tabular}{@{}l@{}}
{\textit{Zero shot}}
\end{tabular}}
&    LaGoVAD$^{\dagger}$~\cite{liu2026language} & 81.12 & 74.25 & --- \\
 &   LAVIDA$^{\dagger}$~\cite{lavida2026} & \underline{82.18} & \underline{90.62} & --- \\
    \midrule

    \multirow{8}{*}{%
\begin{tabular}{@{}l@{}}
{\textit{Training-free}}\\{\textit{zero shot}}
\end{tabular}}
 &   LAVAD~\cite{zanella2024harnessing}  & 80.28 & 62.01 & 85.36 \\
&    AnyAnomaly~\cite{ahn2026anyanomaly} & 80.70 & --- & --- \\
 &   EventVAD~\cite{shao2025eventvad}    & 82.03 & 64.04 & 87.51 \\
 &   VADTree~\cite{li2026vadtree}        & 84.74 & 68.85 & 90.55 \\
 &   PANDA~\cite{Yang2025PANDATG}        & 84.89 & 70.16 & --- \\
 &   Flashback~\cite{lee2025flashback}   & 87.29 & 75.13 & 90.54 \\
 &   ASK-Hint~\cite{askhint2025}         & 89.83 & --- & 90.31 \\
 &   \textbf{NOVA (Ours)} & \textbf{\underline{89.86}} & \underline{84.82} & \textbf{\underline{95.07}} \\
    \bottomrule
  \end{tabular}}
\end{table}

\subsection{Ablation Study}
\label{sec:ablation}

Table~\ref{tab:ablation} reports a full-factorial ablation of NA, VNA, and MAS. When NA is disabled, the normal prompts are generated without the normality-aware constraints introduced in Sec.~\ref{sec:prompts}. The baseline disables all three modules; NA enables normality-aware prompt construction, VNA adds the per-video visual normality anchor, and MAS introduces windowed embeddings and motion gating.

%Table~\ref{tab:ablation} reports a full-factorial ablation of NA, VNA, and MAS using the unconstrained LLM-generated prompt base. Here, NA denotes the normality-aware prompt design introduced in Sec.~\ref{sec:prompts}.  When NA is disabled, the normal prompts are the unconstrained LLM-generated ones, without the disambiguation design.

%Among the three individual components, NA provides the largest and most consistent improvement. Relative to the baseline, NA improves UCF AUC by \textbf{3.17 pp}, XD AUC by \textbf{0.78 pp}, and XD AP by \textbf{3.97 pp}, while introducing no additional inference-time computation. VNA alone improves the same metrics by 1.37, 0.57, and 0.11 pp, respectively. In contrast, the effect of MAS is dataset-dependent: it improves UCF AUC by 1.11 pp, but decreases XD AUC and AP by 1.97 and 7.48 pp.

All configurations use the same PE-Core-L14-336 backbone~\cite{meta2025pecore}, scoring hyperparameters, and SAP scoring rule. This controlled setup isolates the effects of the enabled components.

%All configurations use the same PE-Core-L14-336 backbone~\cite{meta2025pecore}, scoring hyperparameters, and SAP-Softmax rules. This controlled setup isolates the effects of the enabled components.  \textbf{Baseline} (all modules disabled) uses unconstrained LLM-generated anomaly and normal prompts, following the competitive scoring of~\cite{lee2025flashback} and~\cite{dong2024clip}. \textbf{NA} enables our normal-prompt generation design,  \textbf{VNA} adds a per-video visual normal anchor $\mathbf{a}_{\mathrm{VNA}}$, and \textbf{MAS} adds windowed embeddings and motion gating. 

% The final row enables all three components.

\begin{table}[t]
  \centering
  \caption{Full-factorial ablation of NA, VNA, and MAS. Results are reported as mean $\pm$ standard deviation over five independently generated prompt banks. 
  % \textbf{Bold} denotes the best result in each column.
  }
  \label{tab:ablation}
  \resizebox{\columnwidth}{!}{%
  \begin{tabular}{ccc ccc}
    \toprule
    NA & VNA & MAS & UCF AUC (\%) & XD AUC (\%) & XD AP (\%) \\
    \midrule
    $\times$ & $\times$ & $\times$ & $85.62 \pm 1.41$ & $93.51 \pm 0.48$ & $80.36 \pm 1.51$ \\
    \checkmark & $\times$ & $\times$ & $88.79 \pm 0.50$ & $94.29 \pm 0.45$ & $84.33 \pm 1.60$ \\
    $\times$ & \checkmark & $\times$ & $86.99 \pm 1.03$ & $94.08 \pm 0.41$ & $80.47 \pm 1.45$ \\
    $\times$ & $\times$ & \checkmark & $86.73 \pm 1.27$ & $91.54 \pm 0.23$ & $72.88 \pm 0.64$ \\
    \checkmark & \checkmark & $\times$ & $89.54 \pm 0.45$ & $94.85 \pm 0.30$ & $84.65 \pm 1.35$ \\
    \checkmark & $\times$ & \checkmark & $89.34 \pm 0.61$ & $91.07 \pm 0.36$ & $72.96 \pm 0.66$ \\
    $\times$ & \checkmark & \checkmark & $87.54 \pm 1.00$ & $94.65 \pm 0.29$ & $82.40 \pm 0.91$ \\
    \checkmark & \checkmark & \checkmark & $\mathbf{89.86 \pm 0.43}$ & $\mathbf{95.07 \pm 0.33}$ & $\mathbf{84.82 \pm 0.56}$ \\
    \bottomrule
  \end{tabular}}
\end{table}

NA provides the largest and most consistent improvement. Relative to the baseline, NA improves UCF AUC by \textbf{3.17 pp}, XD AUC by \textbf{0.78 pp}, and XD AP by \textbf{3.97 pp}, while introducing no additional inference-time computation. VNA alone improves the same metrics by 1.37, 0.57, and 0.11 pp, respectively. In contrast, the effect of MAS is dataset-dependent: it improves UCF AUC by 1.11 pp, but decreases XD AUC and AP by 1.97 and 7.48 pp.

%Compared to the baseline, NA yields the largest single-factor gain, improving UCF AUC by \textbf{3.17 pp} and XD AP by \textbf{3.97 pp} at no inference-time cost, since it requires only offline prompt construction. VNA improves the three metrics on its own by 1.37, 0.57, and 0.11 pp, respectively. MAS is strongly dataset-dependent; it alone raises UCF-Crime by 1.11 pp but lowers XD-Violence AP by 7.48 pp. The same pattern persists once NA is enabled, where VNA continues to help, while MAS without VNA still costs 11.37 pp of XD AP.

The factorial design also reveals clear interactions between the modules. Combining NA and VNA consistently improves over either component alone. On UCF-Crime, their joint gain is 3.92 pp, smaller than the 4.54 pp obtained by summing their individual gains, suggesting partially overlapping contributions. On XD-Violence, their gains are nearly additive: the joint improvements are 1.34 pp in AUC and 4.29 pp in AP, compared with summed individual gains of 1.35 and 4.08 pp, respectively.

VNA also changes the effect of MAS on XD-Violence. With NA enabled, adding MAS without VNA reduces XD AP from 84.33 to 72.96, whereas enabling VNA together with MAS raises it to 84.82. A similar pattern appears without NA, where VNA+MAS reaches 82.40 AP compared with 72.88 for MAS alone. Once VNA is present, however, the additional accuracy gain from MAS is modest: relative to NA+VNA, the full model improves UCF AUC, XD AUC, and XD AP by 0.32, 0.22, and 0.17 pp, respectively.

MAS nevertheless substantially reduces sensitivity to prompt-bank variation on XD-Violence. Across the four configurations without MAS, the standard deviation of XD AP ranges from 1.35 to 1.60, whereas it decreases to 0.56--0.91 when MAS is enabled. Thus, NA is the primary source of accuracy improvement, VNA provides consistent complementary gains, and MAS mainly improves robustness to prompt-bank variation once combined with VNA. Enabling all three components yields the best mean performance on all three metrics.

Additional representative PCA visualizations and qualitative frame-level anomaly score curves are provided in the supplementary material.

% \subsection{Decomposition of Normal-Side and Anomaly-Side Contributions}

\subsection{Analysis}
\label{sec:decomposition}

\textbf{Normal and Anomaly Prompt Analysis.}
While prompts play an important role in VAD, NOVA focuses on normal-side modeling. Table~\ref{tab:decomp} presents a systematic analysis of different decompositions of normal and anomaly prompts.
A preliminary model uses templates containing only class names, providing weak anomaly descriptions and no normal-side reference.
Replacing the text with LLM-generated anomaly descriptions improves UCF-Crime but does not reliably improve XD-Violence AP, suggesting that enriching anomaly-side modeling alone is insufficient. 
Introducing additional normal descriptions and jointly scoring anomaly and normal prompts with SAP-Softmax, even without NA, further improves performance.
Applying NA further strengthens the normal side by removing semantically ambiguous descriptions that are close to anomalies. These results highlight the importance of normal-side modeling in NOVA.

\begin{table}[t]
  \centering
  \caption{Decomposition of normal-side and anomaly-side contributions on UCF-Crime and XD-Violence.}
  \label{tab:decomp}
  \resizebox{\columnwidth}{!}{%
  \begin{tabular}{lccc}
    \toprule
    Config & UCF AUC (\%) & XD AUC (\%) & XD AP (\%) \\
    \midrule
    Template, anomaly classes only & 66.85 & 90.99 & 77.94 \\
    LLM, anomaly classes only & $72.50 \pm 2.09$ & $91.81 \pm 0.71$ & $76.44 \pm 2.37$ \\
    LLM, normal+anomaly (no NA) & $85.62 \pm 1.41$ & $93.51 \pm 0.48$ & $80.36 \pm 1.51$ \\
    LLM, normal+anomaly (NA) & \textbf{88.79 $\pm$ 0.50} & 
    \textbf{94.29 $\pm$ 0.45} & \textbf{84.33 $\pm$ 1.60} \\
    \bottomrule
  \end{tabular}}
\end{table}

\begin{figure}[t]
  \centering
  \includegraphics[width=\linewidth]{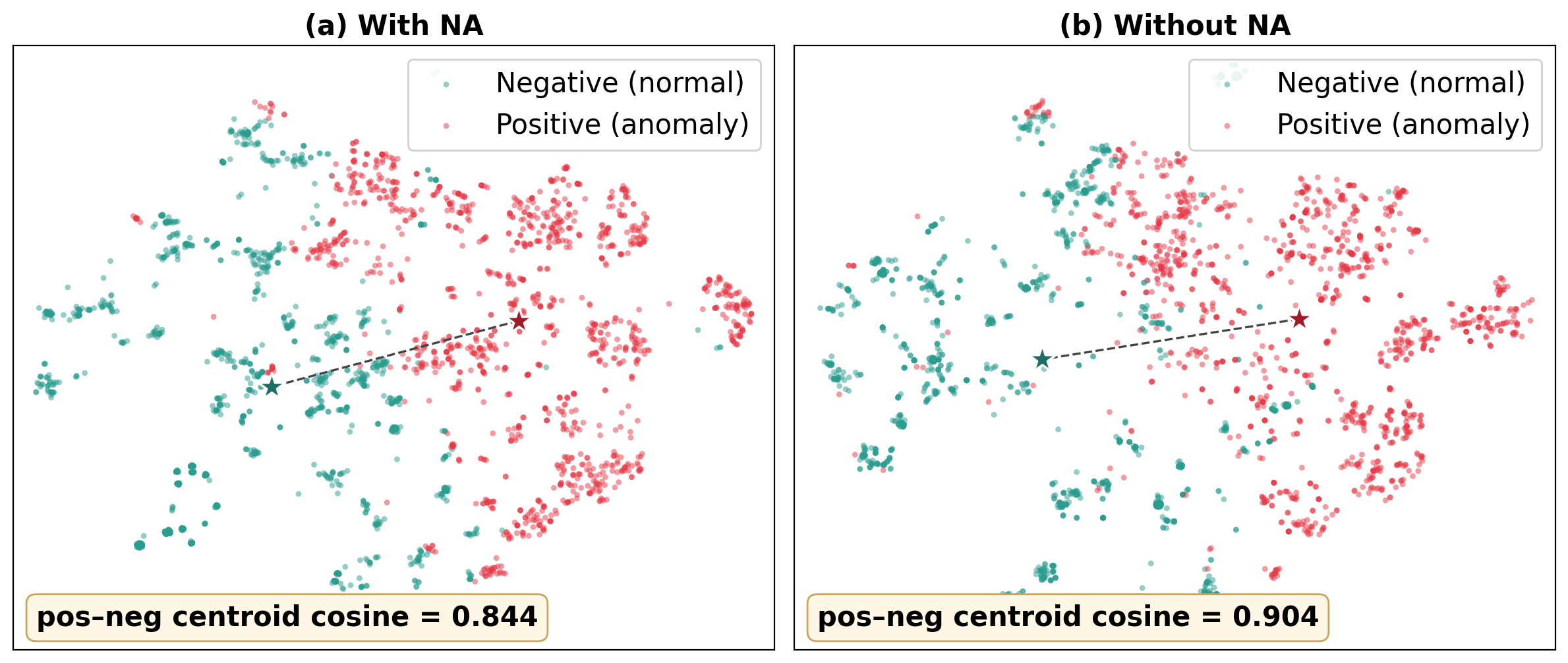}
  \caption{
  t-SNE visualization of UCF-Crime prompt embeddings from five independently generated prompt banks, with and without NA. Stars denote the corresponding centroids; cosine similarities are computed in the original embedding space.
  %\caption{t-SNE projection of UCF-Crime prompt embeddings, where the text embeddings are produced by the PE-Core text encoder over the prompt banks of 5 seeds. Applying NA (left) reduces the anomaly--normal centroid cosine similarity from 0.904 to 0.844, pushing the normal embeddings away from the anomaly descriptions. Stars denote the anomaly/normal centroids. The annotated cosine similarities are computed in the original embedding space.
  % and the annotated value is the cosine similarity between the two centroids, where lower indicates better separation.
  }
  \label{fig:tsne}
\end{figure}

\textbf{Geometric Analysis of NA.}
Fig.~\ref{fig:tsne} visualizes the prompt embeddings before and after applying NA using t-SNE. Compared with unconstrained prompt generation, NA produces a clearer separation between the normal (green) and anomaly (red) prompts, with visibly less overlap. This change is also reflected quantitatively: the anomaly--normal centroid cosine similarity decreases from 0.904 to 0.844. The improved separation is consistent with the ablation results in Table~\ref{tab:ablation}, suggesting that NA strengthens prompt-contrastive scoring by making the normal side more distinguishable from the anomaly side. In contrast, prior work~\cite{dong2024clip} includes normal descriptions with ambiguous verbs such as ``running'' and ``joggers'', which overlap semantically with anomaly-related actions.

\begin{figure}[t]
  \centering
  \includegraphics[width=0.99\linewidth]{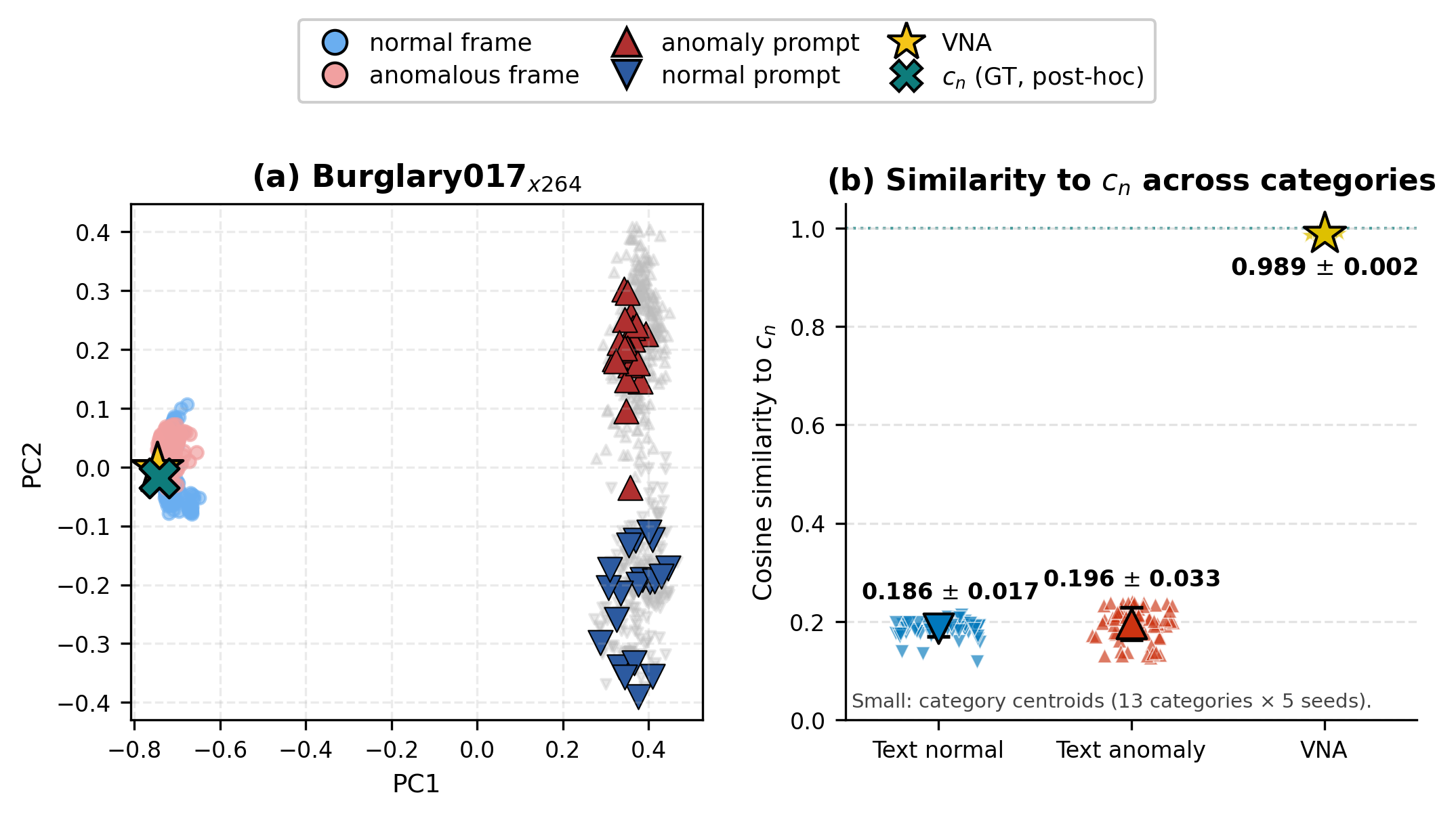}
  \caption{
  Geometric analysis of VNA on the representative UCF-Crime video \emph{Burglary017\_x264}. (a) Per-video PCA projection of frame and prompt embeddings using the seed-10 prompt bank, together with the visual normality anchor $\mathbf{a}_{\mathrm{VNA}}$ and the ground-truth normal-frame centroid $c_n$ used only for post-hoc analysis.
  (b) Original-space cosine similarities to $c_n$ for textual normal and anomaly centroids across 13 anomaly categories and five independently generated prompt banks, together with the five VNA anchors, one per prompt bank. Large markers and error bars denote mean $\pm$ standard deviation.
  %Geometric analysis of VNA on the representative UCF-Crime video \emph{Burglary017\_x264}. (a) Per-video PCA projection using the seed-10 prompt bank. The yellow star denotes the score-weighted visual normality anchor $\mathbf{a}_{\mathrm{VNA}}$, and the teal cross denotes the ground-truth normal-frame centroid $c_n$, which is used only for post-hoc analysis. Blue/red circles are normal/anomalous frames; blue/red triangles are category-specific normal/anomaly prompts; gray markers are prompts from the other categories. (b) Original-space cosine similarities to $c_n$ in three groups: textual normal centroids, textual anomaly centroids, and VNA. Each small text marker represents one category centroid from one prompt bank (13 UCF-Crime categories $\times$ 5 independently generated banks); each small VNA marker represents one prompt bank. Large markers and error bars show mean $\pm$ standard deviation. All three groups share the same continuous vertical scale.
  }
  \label{fig:pca}
\end{figure}

\textbf{Geometric Analysis of VNA.}
\label{sec:vna_geometry}
Fig.~\ref{fig:pca} visualizes the frame embeddings, category prompt embeddings, the visual normality anchor $\mathbf{a}_{\mathrm{VNA}}$, and the visual normal centroid $c_n$ on a per-video PCA plane. Here, $c_n$ is computed only for post-hoc analysis by averaging the normalized embeddings of all ground-truth normal frames and is never used during VNA construction or inference. Panel (a) shows the representative UCF-Crime video \emph{Burglary017\_x264}; panel (b) uses the same visual reference and aggregates textual centroids over all 13 anomaly categories and five independently generated prompt banks. Dataset-level statistics over all 140 anomalous test videos are reported separately below.

As illustrated in Fig.~\ref{fig:pca}, the visual anchor closely matches the visual normal centroid, whereas the textual prompt embeddings remain well separated from the visual frame distribution. Across the 65 category--bank pairs in panel (b), the mean cosine similarity to $c_n$ is $0.186\pm0.017$ for the textual normal centroids and $0.196\pm0.033$ for the textual anomaly centroids. In contrast, the five prompt-bank-dependent VNA anchors for this video achieve $0.989\pm0.002$.

This pattern is consistent across the dataset. Averaged over all videos, the mean cosine similarity between $\mathbf{a}_{\mathrm{VNA}}$ and $c_n$ is 0.969, whereas the corresponding value for the textual normal centroid is only 0.167. Furthermore, in 94.29\% of the videos, the textual anomaly centroid is closer to $c_n$ than the textual normal centroid. These observations support the motivation of VNA: due to the cross-modal gap, textual normal prompts do not necessarily provide a visual reference close to normal frames, whereas the visual anchor constructed from the test video better captures the video's normal appearance. Additional representative examples are provided in the supplementary material.

\section{Conclusions}
\label{sec:conclusion}

We presented NOVA, a training-free ZS-VAD framework that strengthens \textbf{normal-side modeling} in prompt-contrastive anomaly detection. NOVA improves normality representations at the linguistic level through \textbf{Normality-Aware Prompt Construction} and at the visual level through \textbf{Visual Normality Anchor}.
Without training, fine-tuning, or target-data calibration, NOVA achieves 89.86\% AUC on UCF-Crime and 95.07\% AUC / 84.82\% AP on XD-Violence. Ablation and embedding-space analyses further demonstrate the importance of explicit normal-side modeling and the complementary roles of linguistic and visual normality.
These findings suggest that improving anomaly detection does not necessarily require increasingly elaborate anomaly representations; establishing a better boundary between normality and anomaly can provide a stronger basis for anomaly discrimination.

NOVA currently models visual normality from an early-video warm-up prefix and applies the resulting fixed visual anchor throughout the entire video. This design limits its applicability to long-term, open-world video anomaly detection, where normality may evolve over time. These limitations motivate \textbf{online, scene-adaptive normality modeling} as an important direction for future work.

{
    \small
    \bibliographystyle{ieeenat_fullname}
    \bibliography{main}
}

\clearpage
\appendix

\setcounter{page}{1}
\maketitlesupplementary
\setcounter{figure}{0}
\setcounter{table}{0}
\renewcommand{\thefigure}{S\arabic{figure}}
\renewcommand{\thetable}{S\arabic{table}}

\section{Sensitivity to the Prompt-Generation LLM}
\label{sec:llm_sensitivity}

Our framework utilizes an existing LLM to generate normality descriptions.
To assess sensitivity to the prompt-generation LLM, we generate the $M=20$ prompt banks using three representative models: GPT-5.4-mini, Gemini 3.5 Flash
Lite, and Qwen-2.5-72B-Instruct. All prompt-generation roles access these models
through OpenRouter. We use the same generation seed
and keep all downstream evaluation settings fixed. For each prompt bank, we
evaluate both the NA-only configuration and the full NOVA.
%, which additionally enables VNA and MAS.

\begin{table}[t]
  \centering
  \caption{Sensitivity to the prompt-generation LLM. Each model generates $M=20$ positive and 20 negative descriptions per anomaly category. Results are percentages from one prompt-generation seed; all downstream settings are fixed.}
  \footnotesize
  \label{tab:llm_sensitivity}
  \begin{tabular}{llccc}
    \toprule
    \textbf{Generator}  & \textbf{Configuration} & \textbf{UCF AUC} & \textbf{XD AUC} & \textbf{XD AP}  \\
    \midrule
    \multirow{2}{*}{GPT-5.4-mini}     & NA-only     & 88.87 & 92.99 & 79.15 \\
    & Full NOVA & 90.09 & 94.69 & 83.21 \\
        \midrule
    Gemini 3.5  & NA-only   & 89.19 & 92.90 & 79.77 \\
    Flash Lite & Full NOVA & 90.33 & 95.34 & 84.27 \\
             \midrule
    Qwen-2.5-72B- & NA-only  & 87.81 & 94.27 & 84.12 \\
    Instruct & Full NOVA & 89.21 & 95.48 & 85.40 \\
    \bottomrule
  \end{tabular}
\end{table}

As shown in Table~\ref{tab:llm_sensitivity},
across all three generators, the full NOVA consistently improves over the
corresponding NA-only configuration. Moreover, the final results remain within
relatively narrow ranges: 89.21--90.33\% AUC on UCF-Crime, 94.69--95.48\% AUC
on XD-Violence, and 83.21--85.40\% AP on XD-Violence. These results suggest
that NOVA's improvements persist across the evaluated prompt-generation LLMs
and that its final performance is not strongly tied to a particular generator.

% \section{Sensitivity Analysis of \texorpdfstring{$N_\text{warm}$}{Nwarm}}
% \label{sec:nwarm}

\section{Sensitivity to the Number of Prompts}
\label{sec:prompt_count}

We further evaluate how prompt-bank size affects performance. Using the same NA
construction procedure, we generate 100 positive and 100 negative descriptions
per anomaly category. We then take the first $M$ descriptions from each side in
a fixed order, with $M\in\{10,20,40,100\}$, to form prompt banks of different
sizes. Within each dataset, all prompt counts use descriptions from the same
generation run, so the observed differences more directly reflect the impact of $M$.
This experiment uses one prompt-generation seed. NA is enabled, whereas VNA and
MAS are disabled.
The PE-Core encoder, frame sampling, LLM model, SAP-Softmax parameters, and
Gaussian smoothing are otherwise identical to the main evaluation protocol.

\begin{table}[t]
  \centering
  \caption{Sensitivity to the number $M$ of positive and negative prompt descriptions per anomaly category. For each dataset, every row uses the same generation seed and draws $M$ descriptions per side from the same set of 100 positive and 100 negative descriptions. NA is enabled, whereas VNA and MAS are disabled.}
  \small
  \label{tab:prompt_count}
  \begin{tabular}{cccc}
    \toprule
    $M$ & UCF AUC & XD AUC & XD AP \\
    \midrule
    10  & 87.53 & 93.86 & 83.69 \\
    20  & 88.64 & 94.31 & \textbf{84.60} \\
    40  & 89.07 & 94.50 & 83.76 \\
    100 & \textbf{89.22} & \textbf{94.58} & 84.19 \\
    \bottomrule
  \end{tabular}
\end{table}

As shown in Table~\ref{tab:prompt_count}, increasing $M$ consistently improves AUC on both datasets. UCF-Crime AUC rises
from 87.53\% to 89.22\%, while XD-Violence AUC rises from 93.86\% to 94.58\%.
In contrast, XD-Violence AP does not follow the same trend. It reaches its
highest value of 84.60\% at $M=20$, then changes to 83.76\% at $M=40$ and
84.19\% at $M=100$. These results indicate that increasing the number of prompts
can improve AUC without consistently improving AP. The $M=20$ setting used in
the main experiments was fixed before conducting this sensitivity analysis and
was not selected or retuned based on these results. Its performance is
consistent with the observed five-seed range of the original NA-only baseline.

\begin{figure*}[t]
  \centering
  \includegraphics[width=\linewidth]{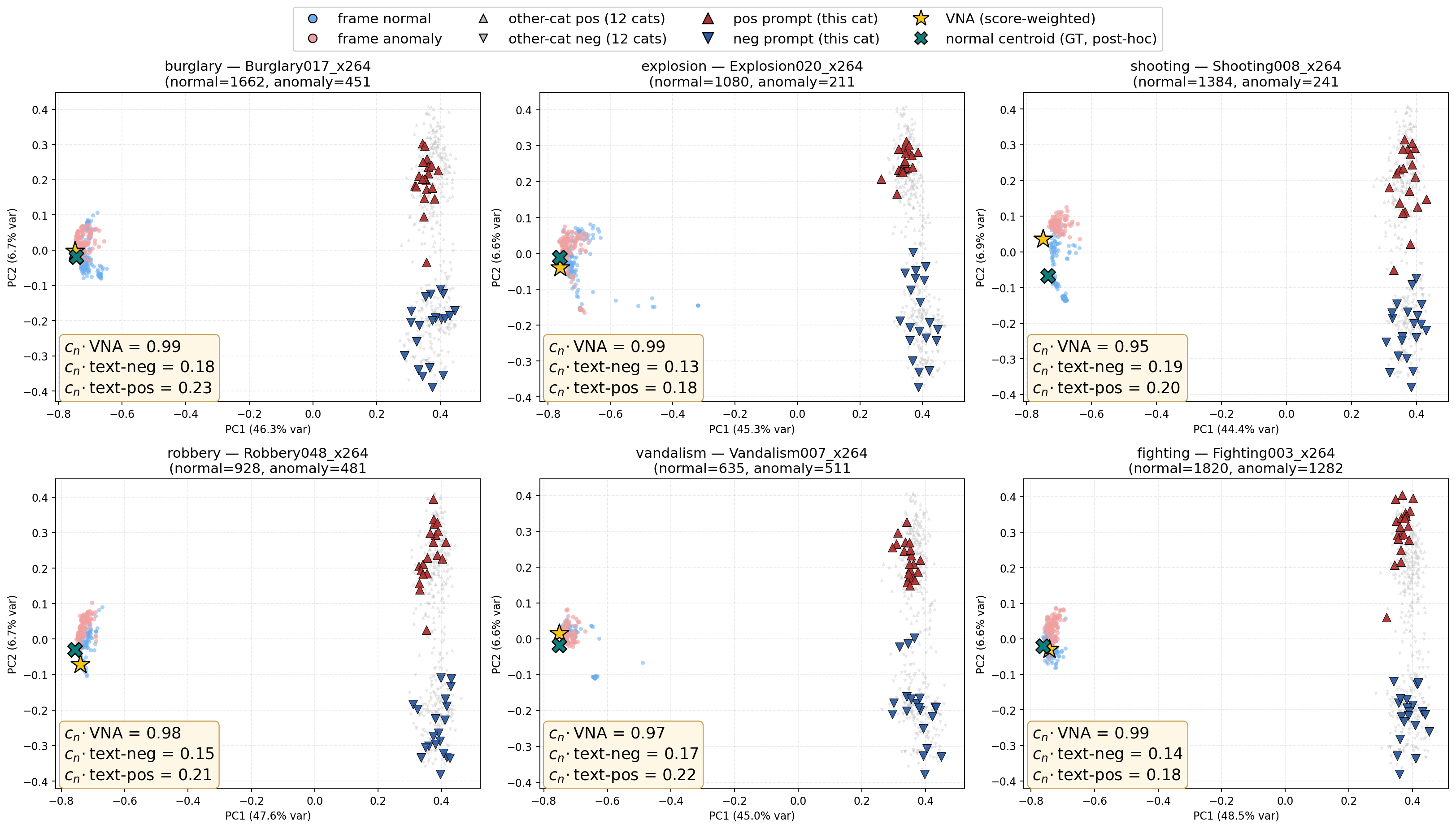}
  \caption{Per-video PCA projections of six illustrative UCF-Crime videos, including the \emph{Burglary017\_x264} example shown in the main paper and five additional examples. The yellow star denotes the score-weighted $\mathbf{a}_{\mathrm{VNA}}$, and the teal cross denotes the ground-truth normal-frame centroid $c_n$. Blue/red dots are normal/anomalous frames; blue/red triangles are category-specific normal/anomaly prompts; gray markers are prompts from the other categories. Annotated cosine similarities are computed in the original embedding space.
  }
  \label{fig:pca_supp}
\end{figure*}

\section{Sensitivity to the Warm-up Length \texorpdfstring{$N_\text{warm}$}{Nwarm}}

Table~\ref{tab:nwarm} evaluates the sensitivity of NOVA to the warm-up window size $N_\text{warm}$ used for constructing the visual normality anchor. Across an eightfold range from 8 to 64 sampled visual embeddings, the performance remains stable on both datasets. The maximum variation is only 0.39 pp in UCF-Crime AUC, 0.57 pp in XD-Violence AUC, and 1.67 pp in XD-Violence AP.

Although performance on XD-Violence increases slightly as $N_\text{warm}$ becomes larger, the improvement is gradual rather than critical, indicating that VNA is not highly sensitive to the exact choice of the warm-up window. We therefore use a fixed value of $N_\text{warm}=32$ throughout all experiments, without dataset-specific tuning.

\begin{table}[t]
  \centering
  \caption{Sensitivity analysis of $N_\text{warm}$. Results are the mean and standard deviation over five prompt banks. $^{\dagger}$ denotes the default used in the main experiments.}
  \small
  \label{tab:nwarm}
  \begin{tabular}{cccc}
    \toprule
    $N_\text{warm}$ & UCF AUC & XD AUC & XD AP \\
    \midrule
    8  & $89.76 \pm 0.40$ & $94.62 \pm 0.38$ & $83.53 \pm 0.68$ \\
    16 & $89.88 \pm 0.41$ & $94.91 \pm 0.35$ & $84.28 \pm 0.59$ \\
    24 & $89.87 \pm 0.42$ & $95.01 \pm 0.33$ & $84.60 \pm 0.58$ \\
    $32^{\dagger}$ & $89.86 \pm 0.43$ & $95.07 \pm 0.33$ & $84.82 \pm 0.56$ \\
    48 & $89.72 \pm 0.45$ & $95.11 \pm 0.33$ & $84.98 \pm 0.54$ \\
    64 & $89.49 \pm 0.47$ & $95.19 \pm 0.33$ & $85.20 \pm 0.53$ \\
    \bottomrule
  \end{tabular}
\end{table}

\section{Additional Geometric Analysis}

Fig.~\ref{fig:pca_supp} reproduces the main-paper \emph{Burglary017\_x264} example alongside five additional illustrative UCF-Crime videos. In each example, the VNA anchor lies substantially closer to the post-hoc visual normal centroid than the category-specific textual normal centroid. These single-prompt-bank visualizations are consistent with the dataset-level statistics reported in the main paper Sec.~\ref{sec:vna_geometry}.

\section{Qualitative Analysis}
\label{sec:qualitative}

\begin{figure}[t]
  \centering
  \includegraphics[width=\linewidth]{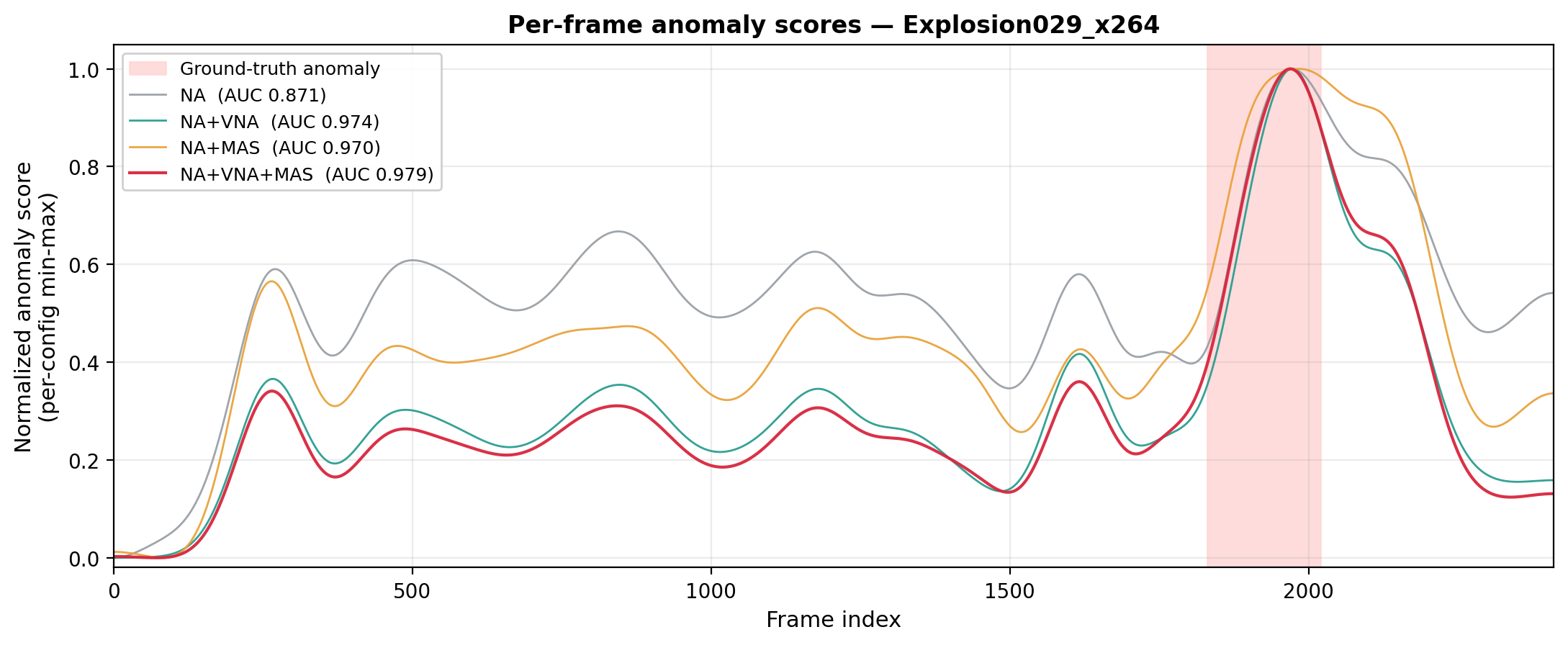}
  \includegraphics[width=\linewidth]{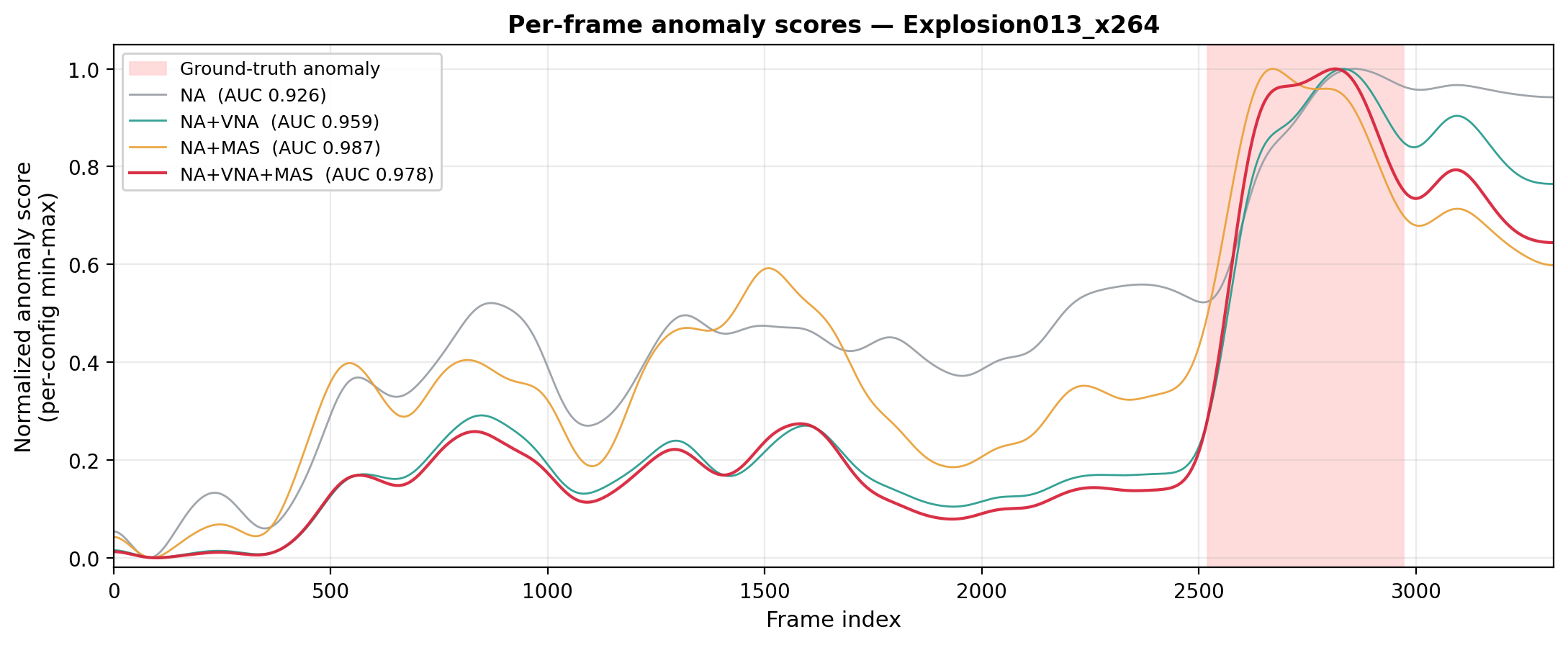}
  \includegraphics[width=\linewidth]{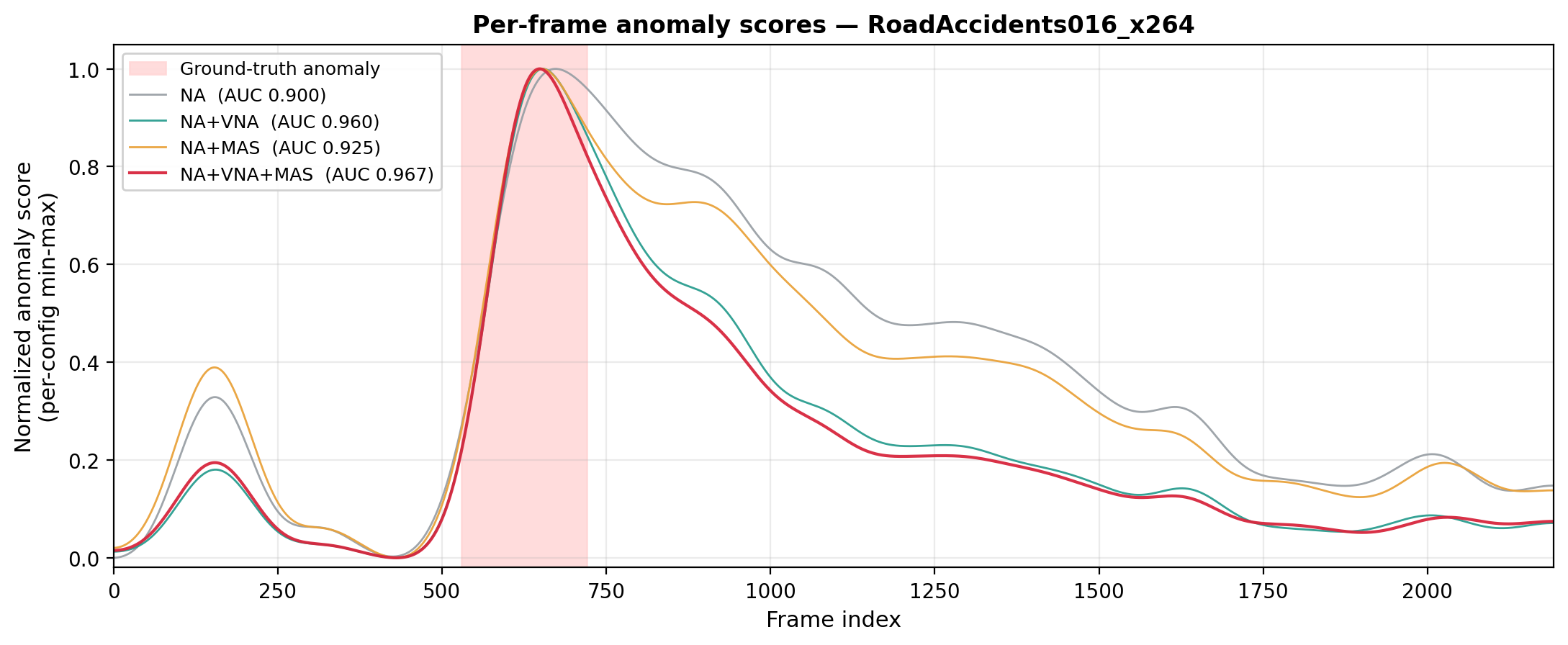}
  \caption{Per-frame qualitative analysis on UCF-Crime. Pink regions denote the annotated ground-truth anomaly frames.
  }
  \label{fig:perframe}
\end{figure}

Fig.~\ref{fig:perframe} shows the frame-level anomaly scores of three representative UCF-Crime videos under different configurations of NOVA. Compared with the baseline, NA generally reduces the anomaly scores assigned to normal segments while retaining clear responses around the annotated anomaly intervals. VNA further suppresses false-positive peaks outside the anomaly regions, leading to a clearer contrast between normal and anomalous frames. MAS mainly improves the temporal smoothness of the score curves by reducing short-lived fluctuations.

\section{Prompt-Generation Templates and Footage Conditioning}
\label{sec:appendix_prompts}

To ensure full reproducibility, this section lists the LLM instruction templates used by each role in the
agentic multi-role pipeline described in main paper Sec.~\ref{sec:prompts}, and provides one
worked example showing how the dataset-level \emph{footage description} shapes the
generated outputs. The manually specified dataset-level context consists of a
footage description together with fixed domain-specific role guidance and examples.
These components are defined once per dataset and reused across categories, seeds,
and prompt-generation runs. The footage description is provided as context to
the prompt-generation agents:

\begin{itemize}
  \item {\bf UCF-Crime:} \texttt{footage = "surveillance footage"}
  \item {\bf XD-Violence:} \texttt{footage = "movie or online video footage"}
\end{itemize}

The remaining placeholders are \texttt{\{anomaly\}} for the current anomaly
category, \texttt{\{M\}} for the number of descriptions per side,
\texttt{\{n\}}/\texttt{\{k\}} for the number of anchors (we use $K=4$),
\texttt{\{fv\}} for the mined confusing actions, and \texttt{\{ex\}} for the
generated calm-anchor examples.

\subsection{Category-Conditioned Confusing Action Mining}
\begin{lstlisting}
[system]
You analyze a CLIP-based anomaly detector for {footage}. For a given anomaly
category you identify AMBIGUOUS actions: benign, normal behaviors that a person
might plausibly do in everyday {footage} but that VISUALLY RESEMBLE the anomaly
and would be confused with it. These are the actions that must be FORBIDDEN from
'normal scene' descriptions so the normal side stays clearly separable from the
anomaly.

[user]
Anomaly category: "{anomaly}".
List 5 to 8 such ambiguous verbs / short action phrases (benign actions that look
like "{anomaly}"). Keep each to 1-3 words, lowercase.
Return ONLY valid JSON: {"verbs": ["...", "..."]}
\end{lstlisting}

\subsection{Calm Anchor Generation}
\begin{lstlisting}
[system]
You write short visual scene descriptions for a CLIP-based anomaly detector. You
produce CALM ANCHOR scenes: ordinary, non-threatening establishing shots typical
of the given footage domain, containing NO conflict, NO weapons and little or no
motion -- the absolute baseline of 'normal' for that domain.

[user]
Footage domain: {footage}.
Write exactly {n} distinct calm anchor scene descriptions (each one short
sentence) typical of this domain.
Return ONLY valid JSON: {"anchors": ["..."]}
\end{lstlisting}

\subsection{Prompt-Generation Skeleton}
The shared generation skeleton is shown below. The \texttt{domain\_intro} field
is shared across datasets, whereas \texttt{positive\_roles} and
\texttt{examples} are supplied by a fixed dataset-specific domain specification.
These fields are specified before prompt generation and then frozen: they are
reused verbatim for every category, every seed, and every run, and no video,
label, or score from the target dataset is used to adapt them.
The \texttt{domain\_intro}, \texttt{positive\_roles}, and \texttt{examples}
fields are identical in the constrained and unconstrained conditions reported
in Table~\ref{tab:ablation} of the main paper. The unconstrained condition
replaces the category-specific normality-aware rules with a generic normal-prompt
instruction, while these shared fields remain unchanged. Their complete
contents are provided below.
\begin{lstlisting}
{domain_intro}

Your task: For the anomaly category "{anomaly}", generate exactly {M} POSITIVES
and {M} NEGATIVES.

--- POSITIVE RULES ---
1. Each describes ONE specific, visually observable atomic action that strongly
   indicates "{anomaly}".
{positive_roles}
3. Include the specific body part, object, or direction of motion.
4. Vary the {M} items across different subjects, stages of the event, and implied
   camera angles.

--- NEGATIVE RULES ---
{negative_rules}

--- HIGH-QUALITY EXAMPLES (use as style reference) ---
{examples}

--- OUTPUT FORMAT ---
Return ONLY valid JSON: {"positives": [ ...{M} ], "negatives": [ ...{M} ]}
\end{lstlisting}

\texttt{domain\_intro} is the same single sentence for both datasets: {\textit{``You are a
Visual Forensic Expert designing text prompts for a CLIP-based video anomaly
detection system.''}} The \texttt{positive\_roles} field names, for a subset of the
categories in $\mathcal{C}$, the kind of subject the description should use, and
closes with the footage description:
\begin{lstlisting}
[UCF-Crime]
2. Use ROLE-SPECIFIC subjects that fit the anomaly:
   - For arrest: "Police officer", "Officer", "Uniformed officer", "Suspect"
   - For fighting: "Person", "Individual", "Attacker", "Victim", "Two people"
   - For arson: "Person", "Individual", "Flames", "Fire"
   - For road accident: "Vehicle", "Car", "Motorcycle rider", "Pedestrian"
   - Match subject to what would realistically appear in surveillance footage.

[XD-Violence]
2. Use ROLE-SPECIFIC subjects that fit the anomaly:
   - For fighting: "Person", "Two people", "Attacker", "Victim", "Brawlers"
   - For shooting: "Gunman", "Shooter", "Armed person", "Victim"
   - For explosion: "Fireball", "Blast", "Debris", "Shockwave"
   - For riot: "Crowd", "Rioters", "Mob", "Clashing protesters"
   - For abuse: "Aggressor", "Victim", "Person"
   - For car accident: "Vehicle", "Car", "Speeding car", "Motorcycle"
   - Match subject to what would realistically appear in movie or online
     video footage of violence.
\end{lstlisting}

The \texttt{examples} field supplies a style reference drawn from two categories
of the corresponding $\mathcal{C}$; it is never used as a verbatim template:
\begin{lstlisting}
[UCF-Crime]
For "abuse" POSITIVES:
"Person forcefully grabbing another by the collar"
"Aggressor twisting victim's arm behind their back forcefully"
"Person slamming another's head against a hard surface"
"Individual restraining victim by sitting on top of them"

For "abuse" NEGATIVES:
"Person sitting quietly on a chair reading"
"Individual standing still near a wall"
"Empty room with furniture and no people"          <- Empty Anchor
"Empty corridor with closed doors on both sides"   <- Empty Anchor
"Person slowly turning around and walking away"

For "arrest" POSITIVES:
"Police officer pressing suspect face-down onto the pavement"
"Officer snapping handcuffs onto a person's wrists behind their back"
"Uniformed officer tackling a fleeing suspect to the ground"
"Officer placing knee on suspect's back while cuffing them"

[XD-Violence]
For "shooting" POSITIVES:
"Gunman aiming a handgun at a fleeing victim"
"Person firing a rifle into a panicking crowd"
"Shooter holding a pistol with both arms extended"
"Muzzle flash from a gun fired in a dark alley"

For "abuse" NEGATIVES:
(same five items as the UCF-Crime block above)
\end{lstlisting}

The \texttt{negative\_rules} block contains the confusing actions
\texttt{\{fv\}} mined by the Verb Miner and the calm-anchor examples
\texttt{\{ex\}} produced by the Anchor Generator. This block implements NA at
the language-content level:
\begin{lstlisting}
1. Must describe SAFE, STATIC, or SLOW behaviors -- the "boring" baseline of
   {footage}.
2. FORBIDDEN verbs/actions for this category -- they visually resemble
   "{anomaly}" and must NEVER appear in a normal description: {fv}.
3. REQUIRED: prefer passive, low-motion, static actions (the calm baseline);
   avoid energetic or fast actions.
4. MANDATORY -- include exactly {k} "Calm Anchor" items: ordinary
   non-threatening establishing scenes of {footage} with NO conflict and NO
   weapons. Examples: {ex}
5. The remaining {M}-{k} items should show people in ordinary, non-threatening
   situations.
6. Vary the subjects and implied camera viewpoints across the items.
\end{lstlisting}

For the unconstrained condition, corresponding to the ``LLM normal+anomaly
(no NA)'' row in Table~\ref{tab:ablation}, the category-specific
confusion-action list, static-action preference, and calm-anchor constraint are
omitted. The generic normal-prompt instructions used in this condition are
shown below.
\begin{lstlisting}
1. Must describe SAFE or non-threatening behaviors -- the normal baseline of
   surveillance footage.
2. The {M} items should show people doing mundane, everyday things.
3. Subject variety: "Person", "Individual", "Two people", "Pedestrian",
   "Shopper", etc.
\end{lstlisting}

\subsection{Geometric Critic and Refinement}
The Geometric Critic does not use an LLM instruction. It embeds each normal
description with the frozen PE-Core text encoder, computes its cosine similarity
$g_j^c$ to the positive centroid, and flags descriptions with
$g_j^c>\theta$ (we use $\theta=0.70$). Flagged descriptions are passed to the
Refiner with the system instruction below. A rewrite is accepted only if it
reduces the ambiguity score, and this procedure is repeated for at most $R=3$
rounds:
\begin{lstlisting}
[system]
You rewrite text prompts for a CLIP-based surveillance anomaly detector. A
'normal scene' description has been flagged because its visual embedding sits too
close to the anomaly '{anomaly}' -- it is visually ambiguous and would be
confused with the anomaly. Rewrite it into an UNAMBIGUOUS normal surveillance
description of the SAME scene type: a plausible benign baseline, but clearly
static / calm / non-threatening so it cannot be confused with '{anomaly}'.
Prefer passive or slow actions (sitting, standing, waiting, walking slowly, empty
scene). Return ONLY the rewritten sentence, no quotes, no explanation.
\end{lstlisting}

\subsection{Effect of Footage Conditioning}
The following example shows prompts generated for the same anomaly category,
\emph{shooting}, under seed 10, with the two datasets' domain specifications.
The \emph{normal} side is conditioned by the footage description at generation
time.

\paragraph{{\bf UCF-Crime.}}
(\texttt{footage = surveillance footage})

\noindent\textit{Positives:}
\begin{itemize}[leftmargin=1.5em,itemsep=0pt,topsep=1pt]
  \item Person extending both arms while firing a handgun toward the left side of the frame
  \item Individual crouched behind a car door with a pistol discharging bright muzzle flashes
  \item Suspect leaning out of a vehicle window while shooting a handgun downward
\end{itemize}
\textit{Negatives:}
\begin{itemize}[leftmargin=1.5em,itemsep=0pt,topsep=1pt]
  \item A quiet empty hallway under fluorescent lights with closed doors and no people.
  \item A static view of a lobby with a reception desk and a few chairs, everything still.
  \item A parking lot at night with parked cars and no visible movement.
\end{itemize}

\paragraph{XD-Violence.} (\texttt{footage = movie or online video footage})

\noindent\textit{Positives:}
\begin{itemize}[leftmargin=1.5em,itemsep=0pt,topsep=1pt]
  \item Gunman extending one arm forward while aiming a handgun at a victim
  \item Shooter firing a pistol from behind a car door toward the street
  \item Armed person gripping a rifle with both hands and pointing it at a doorway
\end{itemize}
\textit{Negatives:}
\begin{itemize}[leftmargin=1.5em,itemsep=0pt,topsep=1pt]
  \item A quiet indoor shot of a person sitting on a couch watching a screen.
  \item A steady view of a desk with a laptop and a mug in a simple room.
  \item A static shot of a living room with soft lighting and a television in the background.
\end{itemize}

\end{document}